\documentclass[review]{elsarticle}

\usepackage{color}
\usepackage{algorithm}
\usepackage{algorithmicx}

\usepackage[noend]{algpseudocode}
\usepackage{amsmath}
\usepackage{amssymb}
\usepackage{multirow}

\usepackage{graphicx}
\usepackage{subfigure}
\usepackage{setspace}
\usepackage{times}
\usepackage{helvet}
\usepackage{courier}

\journal{Journal of Information Sciences}

\begin{document}

\begin{frontmatter}

\title{Recruitment-imitation Mechanism for Evolutionary Reinforcement Learning}

\author[mymainaddress,mysecondaryaddress]{Shuai L\"u\corref{mycorrespondingauthor}}

\ead{lus@jlu.edu.cn}

\author[mymainaddress,mysecondaryaddress]{Shuai Han}
\author[mymainaddress,mysecondaryaddress]{Wenbo Zhou}
\author[mymainaddress,mysecondaryaddress]{Junwei Zhang}

\cortext[mycorrespondingauthor]{Corresponding author}

\address[mymainaddress]{Key Laboratory of Symbolic Computation and Knowledge Engineering (Jilin University), Ministry of Education, Changchun 130012, China}
\address[mysecondaryaddress]{College of Computer Science and Technology, Jilin University, Changchun 130012, China}

\begin{abstract}
Reinforcement learning, evolutionary algorithms and imitation learning are three principal methods to deal with continuous control tasks. Reinforcement learning is sample efficient, yet sensitive to hyper-parameters setting and needs efficient exploration; Evolutionary algorithms are stable, but with low sample efficiency; Imitation learning is both sample efficient and stable,  however it requires the guidance of expert data. In this paper, we propose Recruitment-imitation Mechanism (RIM) for evolutionary reinforcement learning, a scalable framework that combines advantages of the three methods mentioned above. The core of this framework is a dual-actors and single critic reinforcement learning agent. This agent can recruit high-fitness actors from the population of evolutionary algorithms, which instructs itself to learn from experience replay buffer. At the same time, low-fitness actors in the evolutionary population can imitate behavior patterns of the reinforcement learning agent and improve their adaptability. Reinforcement and imitation learners in this framework can be replaced with any off-policy actor-critic reinforcement learner or data-driven imitation learner. We evaluate RIM on a series of benchmarks for continuous control tasks in Mujoco. The experimental results show that RIM outperforms prior evolutionary or reinforcement learning methods. The performance of RIM's components is significantly better than components of previous evolutionary reinforcement learning algorithm, and the recruitment using soft update enables reinforcement learning agent to learn faster than that using hard update.
\end{abstract}


\end{frontmatter}

\section{Introduction}

An important goal of artificial intelligence is to develop agents with excellent decision-making capabilities in complex and uncertain environments. In recent years, the rapid development of deep neural network enables agents based on reinforcement learning methods to perform well in complex control tasks  \cite{lillicrap2015continuous} \cite{mnih2016asynchronous} \cite{schulman2015trust}. However, reinforcement learning methods cannot always handle reward sparse problems effectively and their parameters are very sensitive to disturbances. Recent studies have shown that evolutionary algorithms are better at dealing with sparse reward environments \cite{salimans2017evolution}, and can be employed as an extensible alternative to reinforcement learning in various tasks. When a specific control task has expert data as a guide, imitative learning can also train agents efficiently. Currently, imitative learning-based methods have been successfully applied to drones \cite{giusti2015machine}, automated driving \cite{codevilla2018end}, and other fields.

These artificial intelligence algorithms have close relationships with principles of biology \cite{zhang2017comparing}. Evolutionary algorithms are inspired by evolution of species.  In each generation of the population, there are plenty of random mutations in different individuals. Individuals with positive mutations will be selected by the environment. In imitative learning, individuals with poor fitness can learn behavior patterns of individuals with good fitness in a supervised learning way. At the same time, the diversity of genes (parameters) is preserved. The biological basis of reinforcement learning is that individuals learn correct or wrong actions through trials and errors during interactions with the environment. Off-policy reinforcement learning approaches allow individuals to learn and update their policies from historical data \cite{mnih2015human} \cite{lillicrap2015continuous}. If reinforcement learning individuals are allowed to learn from the entire historical interactions between population and environment, and regularly copy reinforcement learning individuals into the population to participate in the evolution, both the learning process of reinforcement learning and the evolutionary process of evolutionary algorithms can be accelerated \cite{khadka2018evolution} \cite{pourchot2018cem}. However, there are still two problems. 
\begin{itemize}
	\item Reinforcement learning agents can only learn from experience, but not directly accept the guidance of elites in the current population..
	\item Reinforcement learning agents must have the same structure as individuals in the population, or at least have a similar parameterized form. Otherwise, reinforcement learning individuals that are copied into the population may not be able to participate in the evolutionary process.
\end{itemize}

\begin{figure}[!tb]
	\begin{spacing}{0.7}
		\begin{minipage}[t]{1\linewidth}
			\centering
			\includegraphics[width=3.1in]{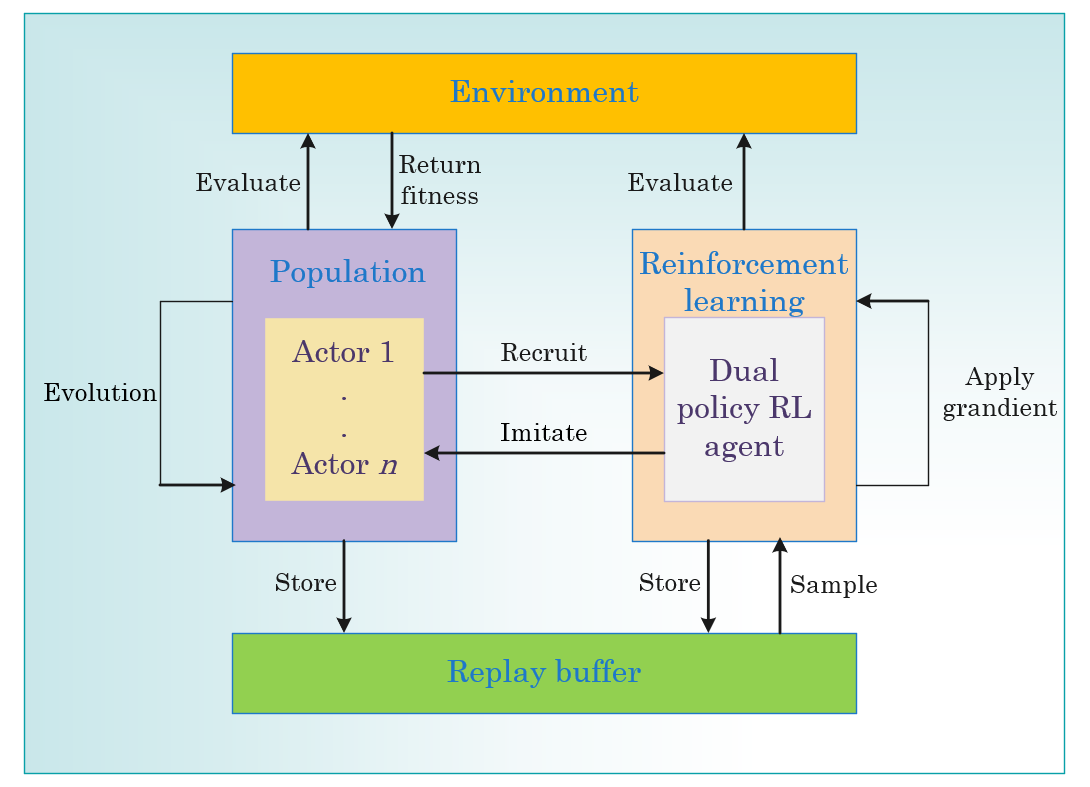}
		\end{minipage}%
		\caption{Framework of RIM for evolutionary reinforcement learning}
		\label{fig:framework}	
	\end{spacing}
\end{figure}

\begin{figure}[!tb]
	\begin{spacing}{0.7}	
		\begin{minipage}[t]{1\linewidth}
			\centering
			\includegraphics[width=3.1in]{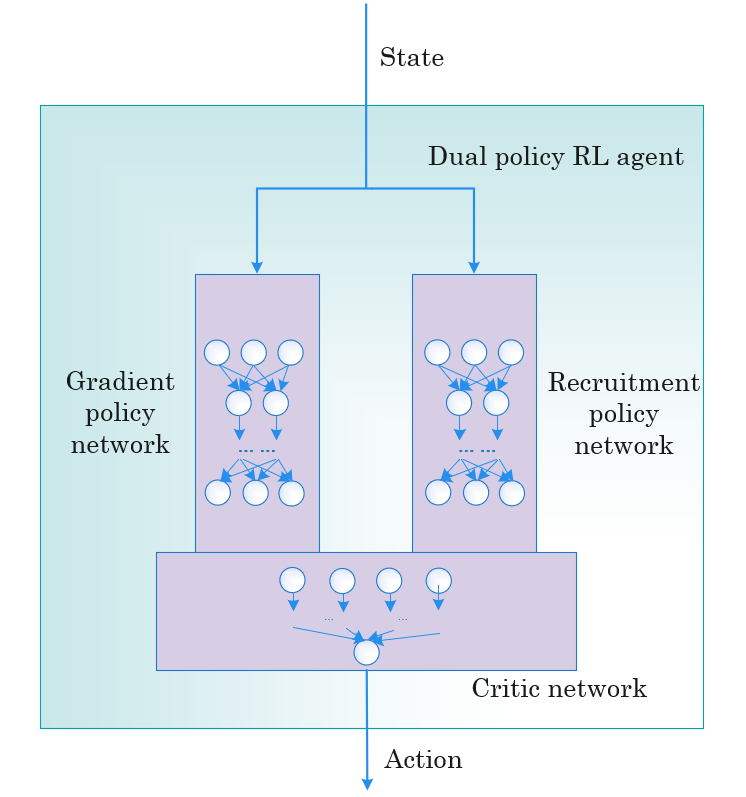}
		\end{minipage}%
		\caption{Structure of dual policy RL agent}	
		\label{fig:rl-agent}
	\end{spacing}
\end{figure}

In response to the above problems, this paper proposes Recruitment-imitation Mechanism (RIM) for evolutionary reinforcement learning. The framework of RIM for evolutionary reinforcement learning is presented in Figure \ref{fig:framework}. 
The recruitment process allows reinforcement learning (RL) agent to recruit the best individual from population to participate in RL agent decision and learning process. This requires RL agent to have the structure of Figure \ref{fig:rl-agent}. In Figure \ref{fig:rl-agent},  gradient policy network and recruitment policy network simultaneously accept the current state as input, and respectively produce actions as output. Critic network compares the potential rewards of actions produced by the two policy networks, and outputs the action with higher reward. The dual-policy decision-making mechanism not only enables the RL agent to produce better experiences, but also enables the RL agent to better estimate $Q$ value, which enables the learning process of RL agent to directly accept the guidance of excellent policy in the population. The imitation process permits low-fitness individuals in the population to learn behavioral patterns of RL agent. Since the structure of RL agent is inconsistent with that of individuals in the population, RL agent cannot be directly injected into the population and participate in the evolution. The imitation process is designed for addressing this problem.

The main contributions of this paper are as follows:
\begin{itemize}
	\item We propose Recruitment-imitation Mechanism (RIM) and an evolutionary reinforcement learning framework that applies this mechanism. At the core of RIM, a dual-policy RL structure is designed, which allows critic network to determines actions.
	\item We present a series of optimization techniques for RIM, including an off-policy imitation learning algorithm that directly uses experience replay buffer and soft updating strategies for recruiting networks.
	\item We compare the performance of RIM with that of previous algorithms on Mujoco benchmarks. Moreover, we discuss the impact of optimization techniques on RIM’s performance.
\end{itemize}

The rest of this paper is organized as follows. Section 2 briefly introduces the background. Section 3 presents the proposed recruitment-imitation mechanism in detail. Section 4 shows the experimental results and comparisons. Section 5 discusses advantages of RIM and outlines the future work.

\section{Related Work}
Combining reinforcement learning with imitation learning or evolutionary algorithms is not a new idea \cite{ross2014reinforcement} \cite{uchibe2018cooperative} \cite{vargas2018evolutionary} \cite{drugan2019reinforcement}. The traditional combination methods mostly consider imitation learning or evolutionary algorithms as a sub-process of reinforcement learning. Typical practices in such methods include: leveraging the distribution estimation method to improve the exploration noise in reinforcement learning \cite{tan2014integration}, utilizing the evolutionary method to optimize the approximation function in $Q$ learning \cite{whiteson2006evolutionary}, or using the imitation learning to optimize the initial stage of the RL process \cite{kober2010imitation}. Unlike these traditional practices, evolution and imitation learning in RIM are not sub-processes of reinforcement learning. In RIM, evolution and reinforcement learning are two relatively independent learning processes, while imitation learning is a way of synchronizing the behavioral policy of dual-policy agents into populations.

Evolutionary Reinforcement Learning (ERL) \cite{khadka2018evolution} provides a new paradigm for the combination of evolutionary algorithms and reinforcement learning. ERL's approach is to reuse the interaction data between the population and the environment, and inject RL policy into the population. RIM can be viewed as an extension of ERL. The expanded parts include: 

1) Combining evolutionary algorithms from the perspective of reinforcement learning. Namely, recruiting policy directly from the population to participate in reinforcement learning processes; 

2) Constructing a dual-policy agent in the RL component. The agent can integrate two policies to generate experience, and give better estimation of $Q$ value when policy learning is insufficient; 

3) Leveraging off-policy imitation learning to synchronize the behavioral policy of dual-policy agent into the population. By the way, the RL agent does not have to maintain the same structure as the individuals in the population.

In the latest work, there are other ways to extend ERL: Collaborative Evolutionary Reinforcement Learning (CERL) \cite{khadka2019collaborative} uses different combinations of learners to complete exploration in different time-horizons of tasks; Cross-entropy Method Reinforcement Learning (CEM-RL) \cite{pourchot2018cem} replaces the combination of standard evolutionary algorithm and DDPG \cite{lillicrap2015continuous} in ERL with a combination of Cross-entropy Method (CEM) and Twin Delayed Deep Deterministic policy gradient (TD3) \cite{fujimoto2018addressing}. Unlike the above works, RIM is neither an optimization for task exploration nor a replacement of sub-components in ERL. Instead, it introduces a recruitment imitation mechanism that enables reinforcement learning, imitation learning, and evolutionary algorithms to be more effectively combined.  In other words, RIM, CERL and CEM-RL are independent optimizations in different directions of ERL.

\section{Background}
This section introduces the background of deep reinforcement learning, evolutionary algorithms and imitation learning.
\subsection{Deep reinforcement learning}
Reinforcement learning methods abstract the interaction between an agent and environments into a Markov decision process. At each discrete time step $t$, the environment provides an observation \begin{math} s_t  \end{math} to the agent, and the agent takes an action \begin{math} a_t \end{math} as a response to this observation. Then, the environment returns a reward $r_t$ and the next state \begin{math} s_{t+1} \end{math} to the agent. The goal of reinforcement learning is to maximize the expectation of \begin{math} R_t = \sum_{k=0}^{K}\gamma^kr_{t+k} \end{math}, where \begin{math} R_t \end{math} indicates the sum of cumulative discounted rewards of $K$ steps from the current moment. \begin{math} \gamma \in (0,1] \end{math} is a discount factor.

Deep deterministic policy gradient (DDPG) is a widely used model-free reinforcement learning algorithm based on actor-critic structure \cite{lillicrap2015continuous}. In DDPG,  actor and critic are parameterized as \begin{math} \pi(s|\theta^{\pi}) \end{math} and \begin{math} Q(s,a|\theta^Q) \end{math} respectively. They are called current networks. In addition, replications of actor and critic networks \begin{math} \pi'(s|\theta^{\pi'}) \end{math} and \begin{math} Q'(s,a|\theta^{Q'}) \end{math} are used to provide consistent targets for the learning process. They are called target networks. During the learning process, target networks will be soft updated based on current networks and a weighting factor $\tau$. DDPG completes off-policy learning by sampling experiences from a repaly buffer $B$. That is, for each interaction between an agent and environments, the tuple $(s_t, a_t, r_t, s_{t+1})$ is stored into the replay buffer. One advantage of off-policy reinforcement learning is the sample efficiency of these algorithms. The other advantage is that learning from historical data can decouple exploration process and learning process. DDPG's exploration policy is constructed from actor policy and noise: $\pi_e(s_t)=\pi(s_t|\theta^{\pi})+\mathcal{N}$. The actor and critic networks are updated by sampling mini-batch experiences from the replay buffer. Critic network is updated by minimizing the loss function as follows:
$$ L = \frac 1{N} \sum_{i}(y^{(i)}_t-Q(s^{(i)}_t,a^{(i)}_t|\theta^{Q}))^2\eqno(1) $$
where $y^{(i)}_t=r^{(i)}_t+\gamma Q'(s^{(i)}_{t+1},\pi'(s^{(i)}_{t+1}|\theta^{\pi'})|\theta^{Q'})$. The actor network is updated according to sampled policy gradient:
$$ \bigtriangledown_{\theta^{\pi}}\pi|_{s^{(i)}} \approx \frac 1{N}\sum_{i}\bigtriangledown_aQ(s^{(i)},a|\theta^Q)|_{a=\pi(s^{(i)})}\bigtriangledown_{\theta^{\pi}}\pi(s^{(i)}|\theta^{\pi}) \eqno(2) $$

\subsection{Evolutionary algorithms}
Evolutionary algorithms (EAs) are black-box optimization algorithms that are inspired by natural evolutionary processes. The evolutionary process acts on a population composed of several parameterized candidate solutions (individuals). During each iteration, parameters of individuals in the population are randomly perturbed (mutated), enabling new individuals to be generated. After generating new individuals, the environment evaluates their fitness. Individuals with higher fitness have higher probability to be selected, and selected individuals will participate in the next iterative process. When evolutionary algorithms are implemented to solve control tasks, several actor networks will participate in the evolutionary process. During the iterative process, actor networks with higher fitness will be retained as elites and shielded mutation step \cite{khadka2018evolution}. EAs can be employed as an alternative for reinforcement learning techniques based on Markov decision processes \cite{salimans2017evolution}.

\subsection{Imitation learning}
In imitation learning, an actor imitates expert behaviors to obtain a policy that is close to expert performance. Dataset aggregation (DAgger) method \cite{ross2011no} is a widely used imitation learning algorithm. It is an iterative reduction to online policy training method. Suppose that there is an expert data set $D=\{s_1,a_1,...,s_n,a_n\}$. During each iteration, the actor is trained on this dataset in a supervised manner. The loss function during training can be denoted as: $J(\theta) = \frac 1{K}\sum_{k=1}^{K}L(\pi_{il}(s_k), a_k)$, where $\pi_{il}$ is a policy to be trained and $K$ is the batch size of samples. After convergence, policy $\pi_{il}$ is carried out to generate a state access set $D_{\pi}=\{s_{n+1},...,s_{n+m}\}$. Then $D_{\pi}$ is labeled by actions output from the expert policy, and the expert data is accumulated: $D\gets D \cup D_{\pi}$. Then proceed to the next iteration. The advantage of DAgger is that it can employ expert policy to teach actors how to recover from errors. DAgger is a class of Follow-The-Leader algorithms \cite{attia2018global}.

\section{Recruitment-imitation Mechanism}
Recruitment-imitation mechanism in evolutionary reinforcement learning is presented in this section.
\subsection{Dual policy reinforcement learning agent}
The core of recruitment-imitation mechanism is a dual policy RL agent with dual-actors and single-critic. In the iterative process of the evolutionary reinforcement learning algorithm, the dual policy RL agent can recruit excellent individuals from the population to participate in decision-making or reinforcement learning process. In addition, individuals with poor performance in the population can periodically imitate behavior patterns of dual policy RL agent to accelerate evolutionary speed of the population.

Dual policy RL agent in recruitment-imitation mechanism still uses the actor-critic structure but has two actor networks, including a gradient policy network $\pi_{pg}(s_t|\theta^{pg})$ and a recruitment policy network $\pi_{ea}(s_t|\theta^{ea})$. When the agent makes decisions, the critic network will identify which policy under the current state $s_t$ is potentially more profitable. Therefore, policy of the dual policy RL agent can be indicated as:

$$
\pi_{rl}=\left\{
\begin{array}{rcl}
\pi_{pg}, & & {Q(s_t, \pi_{pg}(s_t|\theta^{pg})|\theta^Q) \geq Q(s_t, \pi_{ea}(s_t|\theta^{ea})|\theta^Q)}\\\\
\pi_{ea}, & & otherwise
\end{array} \right. \eqno(3)
$$
where $\pi_{rl}$ indicates the overall behavior policy of the dual policy RL agent. Similar to DDPG, during the learning process, the critic network is updated by minimizing the loss function of (1), except that $y^{(i)}_t$ is estimated using $\pi_{rl}$:
$$ y^{(i)}_t=r^{(i)}_t+\gamma Q'(s^{(i)}_{t+1},\pi_{rl}(s^{(i)}_{t+1})|\theta^{Q'}) \eqno(4) $$
The gradient policy network is still updated with the sampled policy gradient according to equation (2).

There are two reasons that the dual policy RL agent can make RIM perform better: 

\begin{itemize}
	\item The dual policy RL agent can generate better experiences according to the behavior pattern in equation (3). 
	\item When gradient update is performed, $\pi_{rl}$ used in equation (4) can make more accurate estimation of $Q'$. 
\end{itemize}

Then we prove that equation (4) can give more accurate estimation of $Q'$.

\newtheorem{thm}{\bf Theorem}
\begin{thm}\label{thm1}
	Suppose that $Q'$ and $Q$ converge at the same position for policy $\pi^*$. The converged policy to maximize $Q'$ or Q is $\pi^*$, and the estimation value using $\pi^*$ is $Q^*$. The mean of $Q'$ estimation of DDPG is denoted as $E_{ddpg}(\hat{Q'})$ and the mean of $Q'$ estimation of RIM is denoted as $E_{rim}(\hat{Q'})$. Then, $E_{ddpg}(\hat{Q'}) \leq E_{rim}(\hat{Q'}) \leq Q^*$.
\end{thm}

{{\noindent\it Proof:}\quad}
{\it Since $\pi^*$ is the policy that maximizes $Q'$ (i.e., $Q^*$), any policy that is not $\pi^*$ will cause the estimate of $Q'$ to be less than $Q^*$. So, there is $E_{rim}(\hat{Q'}) \leq Q^*$. $E_{rim}(\hat{Q'}) = Q^*$ if and only if the policy of $\pi_{rl}$ in equation (3) is always $\pi^*$. Also there is $E_{ddpg}(\hat{Q'}) \leq Q^*$. $E_{ddpg}(\hat{Q'}) = Q^*$ if and only if the policy $\pi_{ddpg}$ of DDPG is always $\pi^*$ .
	
	Assume that DDPG's behavior policy $\pi_{ddpg}$ causes the estimation of $Q'$ to be shifted downward by $x$ with the probability of $p(x)$, then $E_{ddpg}(\hat{Q'})=\int_{-\infty}^{+\infty}(Q^*-x)p(x)dx$. Since RIM uses the policy in equation (3) to estimate $Q'$ in equation (4). When $\pi_{pg}$ (i.e., $\pi_{ddpg}$) estimates $Q'$ downwards by $x$ with the probability of $p(x)$, $\pi_{ea}$ can upwards correct $y$ ($y\geq0$) with the probability of $q(y|x)$. Then we have:
	
	\begin{spacing}{1.0}
		\begin{small}
			\begin{align*}
			E_{rim}(\hat{Q'})&=\int_{-\infty}^{+\infty}\int_{-\infty}^{+\infty}(Q^*-x+y)q(y|x)p(x)dydx\\
			&=\int_{-\infty}^{+\infty}\int_{-\infty}^{+\infty}(Q^*-x)q(y|x)p(x)dydx +\int_{-\infty}^{+\infty}\int_{-\infty}^{+\infty}yq(y|x)p(x)dydx\\
			&=\int_{-\infty}^{+\infty}(Q^*-x)(\int_{-\infty}^{+\infty}q(y|x)dy)p(x)dx +\int_{-\infty}^{+\infty}\int_{-\infty}^{+\infty}yq(y|x)p(x)dydx\\
			&=\int_{-\infty}^{+\infty}(Q^*-x)p(x)dx +\int_{-\infty}^{+\infty}\int_{-\infty}^{+\infty}yq(y|x)p(x)dydx\\
			&=E_{ddpg}(\hat{Q'})+\int_{-\infty}^{+\infty}\int_{-\infty}^{+\infty}yq(y|x)p(x)dydx\\
			\end{align*}
		\end{small}
	\end{spacing}
	
	Considering $y\geq0$ and $q(y|x)p(x)\geq0$, then $\int_{-\infty}^{+\infty}\int_{-\infty}^{+\infty}yq(y|x)p(x)dydx \geq 0$, so it holds that $E_{ddpg}(\hat{Q'})\leq E_{rim}(\hat{Q'})$.$\hfill\blacksquare$ 
	
}
In Theorem 1, we can assume that $Q$ and $Q'$ converge at the same position because $Q$ and $Q'$ are very close using soft update setup in continuous domains \cite{fujimoto2018addressing}. Theorem 1 states that the estimation of $Q'$ using $\pi_{rim}$ is more accurate than that estimated using $\pi_{ddpg}$, or closer to $Q^*$. Unlike overestimation problems \cite{hasselt2010double} \cite{van2016deep}, Theorem 1 states that when $\pi_{ddpg}$ does not converge sufficiently during the updating process according to equation (2), if the policy $\pi_{ea}$ in population gives better actions, the $Q'$ estimation of RIM will be more accurate. This is one of the reasons the RL component of RIM can learn better.

\subsection{Off-policy imitation learning}

Due to the inconsistencies in structures of a dual policy RL agent (dual-actors and single-critic) and an individual (single-actor) in the population, the dual policy RL agent cannot be directly injected into the population to participate in the evolution. In this case, an ideal solution is to use imitation learning to unify behavior patterns of the RL agent and individuals in the population.

\begin{figure}[!t]
	\begin{spacing}{0.7}	
		\begin{minipage}[t]{1\linewidth}
			\centering
			\includegraphics[width=3in]{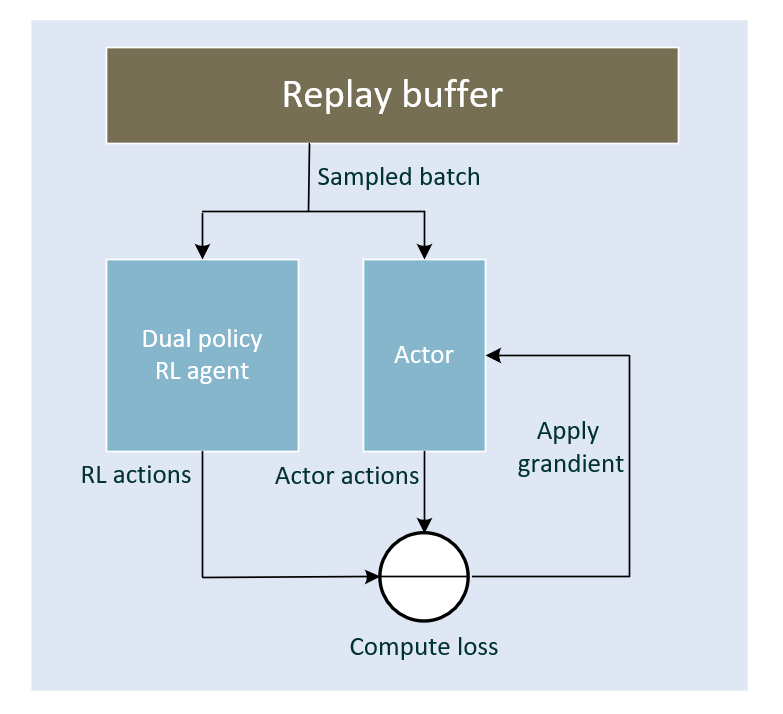}
		\end{minipage}%

		\centering
		\caption{Framework of off-policy imitation learning in RIM.}
		\label{fig:off-policy-il}	
	\end{spacing}
	
\end{figure}

\begin{figure}[!t]
	\begin{spacing}{0.7}	
		\centering
		\subfigure[walker2d-v2]{
			\begin{minipage}[t]{0.5\linewidth}
				\centering
				\includegraphics[width=2.2in]{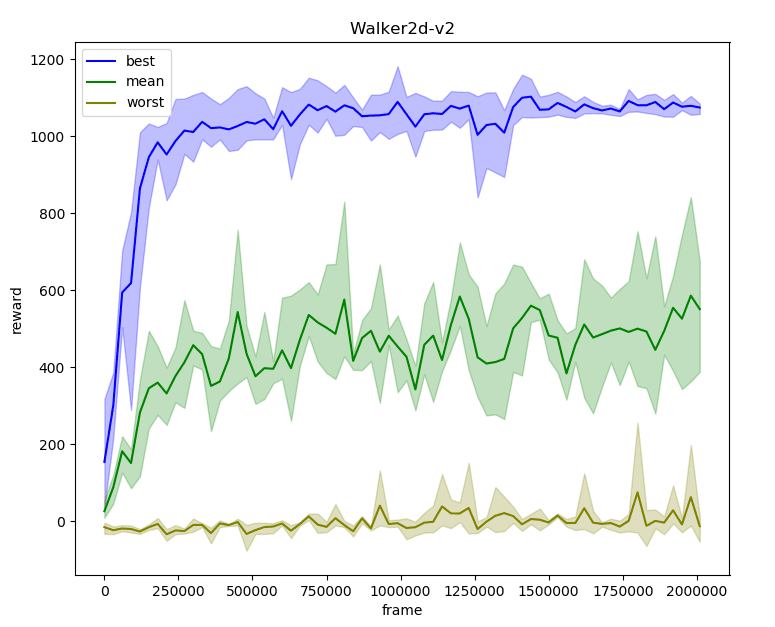}
			\end{minipage}%
			\label{fig:ea_walker2d}
		}%
		\subfigure[Hopper-v2]{
			\begin{minipage}[t]{0.5\linewidth}
				\centering
				\includegraphics[width=2.2in]{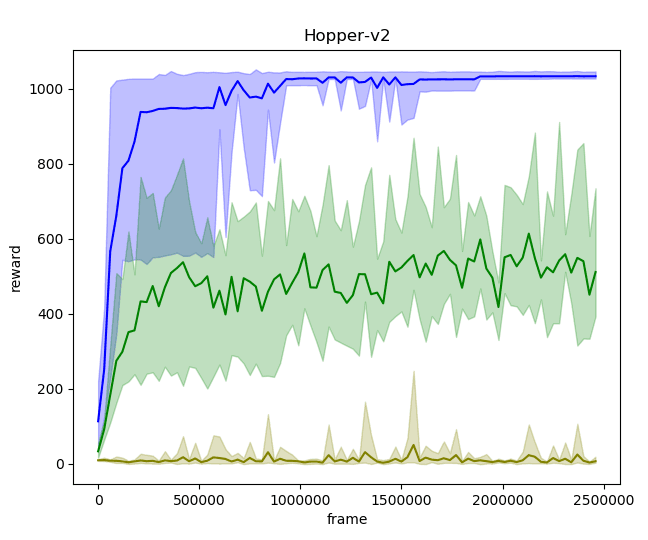}
			\end{minipage}%
			\label{fig:ea_hopper}
		}%
		\caption{Performance of different individuals in the population.}
		\label{fig:ea_m_s}	
	\end{spacing}
\end{figure}

As a typical imitation learning algorithm, DAgger can get better performance with only a few iterations when the amount of data is large. Evolutionary reinforcement learning system usually has a huge experience replay buffer for storing historical data. The actors in the population to be trained can directly sample states and actions in the experience replay buffer and be trained according to the following loss function:

$$ J_0(\theta^{wt}) = \frac 1{K}\sum_{k=1}^{K}L(\pi_{wt}(s_k|\theta^{wt}), \pi_{rl}(s_k)) \eqno(5) $$
where $\pi_{wt}$ indicates the policy of an actor to be trained. $\pi_{wt}$ is usually a policy of an actor with the worst performance in the population. Figure \ref{fig:off-policy-il} shows the framework of off-policy imitation learning and Algorithm \ref{alg:offpolicy-il} describes the detailed process of off-policy imitation learning in RIM. \begin{algorithm}[!t]
	\caption{Off-policy imitation learning in RIM}
	
	\begin{spacing}{1}
		
		\begin{algorithmic}[1]    
			\State Get the worst policy $\pi_{wt}$ from population
			\For{$n$ = 1 to $N$}     
			\State Get replay buffer size: $ size \gets len(B)$
			\While {$size > l_0$}
			\State Sample random batch $\{s_i\}$ from replay buffer
			\State Get $a_i^{wt}$ according to $\pi_{wt}(s_i|\theta^{wt})$
			\State Get $\pi_{rl}$ according to equation (3)
			\State Get expert action $a_i$ according to $\pi_{rl}(s_i)$
			\State Calculate loss $L$ between expert action $a_i^{wt}$ and agent action $a_i$
			\State Calculate $\bigtriangledown_{\theta^{wt}} L$ 
			\State Update $\theta^{wt} \gets \theta^{wt} + \alpha\cdot\bigtriangledown_{\theta^{wt}} L$ 
			\EndWhile
			\EndFor
		\end{algorithmic}
	\end{spacing}
	\label{alg:offpolicy-il}
\end{algorithm}We cancel the data labeling process and accumulation process of DAgger so that the imitation process is off-policy completely. The main function of DAgger's data labeling process and accumulation process is to recover the imitation learner from errors. In the RIM environment setting, there is a large amount of erroneous data in the experience replay buffer, and most of the states sampled from the experience replay buffer is generated by a suboptimal or wrong policy. As shown in Figure \ref{fig:ea_m_s}, the performance of individuals in the population varies widely, and even the interaction information of the worst individual is stored into the replay buffer. Therefore the training process using equation (5) is also a process to teach actors how to recover from an error state.

\subsection{Learning and evolution process of RIM}

The core idea of RIM is to 1) accelerate the learning process by recruiting elite individuals in the population to guide the policy gradient network learning, and 2) accelerate the evolutionary process by enabling individuals with poor performance in the population to imitate behavioral patterns of the dual policy RL agent. 

The process of RIM is as follows: 

1) Some actor networks in the population and actor and critic networks in the dual policy agent are initialized with random weights. 

2) In the evolution process, the environment evaluates the fitness of actors in the population and select a part of the actors to survive with a probability according to the fitness. 

3) The actors are perturbed by mutation and crossover to generate the next generation of actors. 

4) After generating the next generation, the dual policy RL agent recruits an actor with the highest fitness in the population, copies its parameters to the recruitment policy network, and learns for a certain number of times. 

Algorithm \ref{alg:learning} describes the detailed learning process of dual policy RL agent. The worst actor in the population imitates behavioral patterns of the dual policy RL agent every several generations so that the learning outcomes of the dual policy RL agent can be injected into the population. \begin{algorithm}[!tb]
	\caption{Learning process of RL agent in RIM}
	\begin{algorithmic}[1]
		\State Randomly initialize critic $Q(s,a|\theta^Q)$, RL actor $\pi_{pg}(s|\theta^{pg})$ and EA actor $\pi_{ea}(s|\theta^{ea})$      
		\State Initialize target network $Q'$, $\pi_{pg}'$, $\pi_{ea}'$ with $\theta^{Q'} \gets \theta^{Q}$, $\theta^{pg'} \gets \theta^{pg}$, $\theta^{ea'} \gets \theta^{ea}$
		\State Initialize replay buffer $B$, environment $Env$
		\For{$episode$ = 1 to $M$}     
		\State Recruit champion from populations: $\theta^{ea} \gets \theta^{champ}$
		\State Reset $Env$ and get a state $s$
		\State $ done \gets False$
		\While {$!done$}
		\State $\pi(s) \gets  max_\pi\{Q(s,\pi_{pg}(s)), Q(s, \pi_{ea}(s))\}$
		\State Get action $a$ according to $\pi(s) + \mathcal N$
		\State Execute $a$ and observe $ (r, s', done) \sim Env $
		\State Store experience $ (s,a,r,s') $ into replay buffer $B$
		\State Update state: $ s \gets s'$
		\If{$len(B) > l_0$}
		\State Sample random batch $\{(s_t, a_t, r_t, s_{t+1})_i\}$ from replay buffer
		\State $Q'(s_{t+1}^{(i)},a_{t+1}^{(i)}|\theta^{Q'})\gets max\{Q'(s_{t+1}^{(i)},\pi_{pg}'(s_{t+1}^{(i)})|\theta^{Q'}), Q'(s_{t+1}^{(i)}, \pi_{ea}'(s_{t+1}^{(i)})|\theta^{Q'})\}$
		\State $L^Q \gets\frac{1}{N}\sum_i(r_t^{(i)}+Q'(s_{t+1}^{(i)},a_{t+1}^{(i)}|\theta^{Q'})-Q(s_{t}^{(i)},a_{t}^{(i)}|\theta^{Q}))^2$
		\State $L^{\pi} \gets -\frac{1}{N}{\sum_iQ(s_{t}^{(i)},\pi_{pg}(s_{t}^{(i)}|\theta^{pg})|\theta^{Q})} $
		\State Update $\theta^Q$, $\theta^{pg}$ according to $\bigtriangledown_{\theta^Q} L^Q$, $\bigtriangledown_{\theta^{pg}} L^{\pi}$
		
		\State$ \theta^{ea'} \gets (1-\tau_0)\theta^{ea'} + \tau_0\theta^{ea} $
		\State$ \theta^{Q'} \gets (1-\tau_1)\theta^{Q'} + \tau_1\theta^{Q} $
		\State$ \theta^{pg'} \gets (1-\tau_1)\theta^{pg'} + \tau_1\theta^{pg} $
		\EndIf
		
		\EndWhile
		\EndFor
	\end{algorithmic}
	\label{alg:learning}
\end{algorithm}

During the evolutionary process, interaction information between actors and environments will be stored into the experience replay buffer. These experiences are not only used for the sampling process of gradient policy network learning in Algorithm \ref{alg:learning}, but also for the sampling process of the worst actor imitating in population. In the RIM setting of this paper, the structure of gradient policy network is same as that of actors in the population, so when the imitation learning is not performed, the gradient policy network will be periodically copied into the population to accelerate the evolutionary process.

The recruitment policy network participates in the decision process of $\pi_{rl}$. Therefore, it affects the value of $y^{(i)}_t$. The learning process of critic network needs to be provided with a consistent $y^{(i)}_t$. Drawing on the idea of soft update in DDPG, we soft update the target gradient policy network and critic network. In order to further ensure consistency in the learning process, we decide to add a target recruitment policy network. Current recruitment policy network $\pi_{ea}(s_t|\theta^{ea})$ uses hard update to copy directly from the population, and the target recruitment policy network uses soft update to update parameters:
$$ \theta^{ea'} \gets (1-\tau_0)\theta^{ea'} + \tau_0\theta^{ea} \eqno(6) $$
where $\tau_0$ should be much less than 1. The recruitment process using target recruitment network is called soft update in recruitment, and the recruitment process without target recruitment network is called hard update in recruitment. We will perform comparative experiments on these two recruitment models in the experimental part.





\section{Experiments}
In this section, we show comparative experiments and a series of analytical results.

\subsection{Experimental settings}

\begin{figure*}[!htb]
	\begin{spacing}{0.7}	
		\centering
		\subfigure[Walker2d-v2]{
			\begin{minipage}[t]{0.5\linewidth}
				\centering
				\includegraphics[width=2.1in]{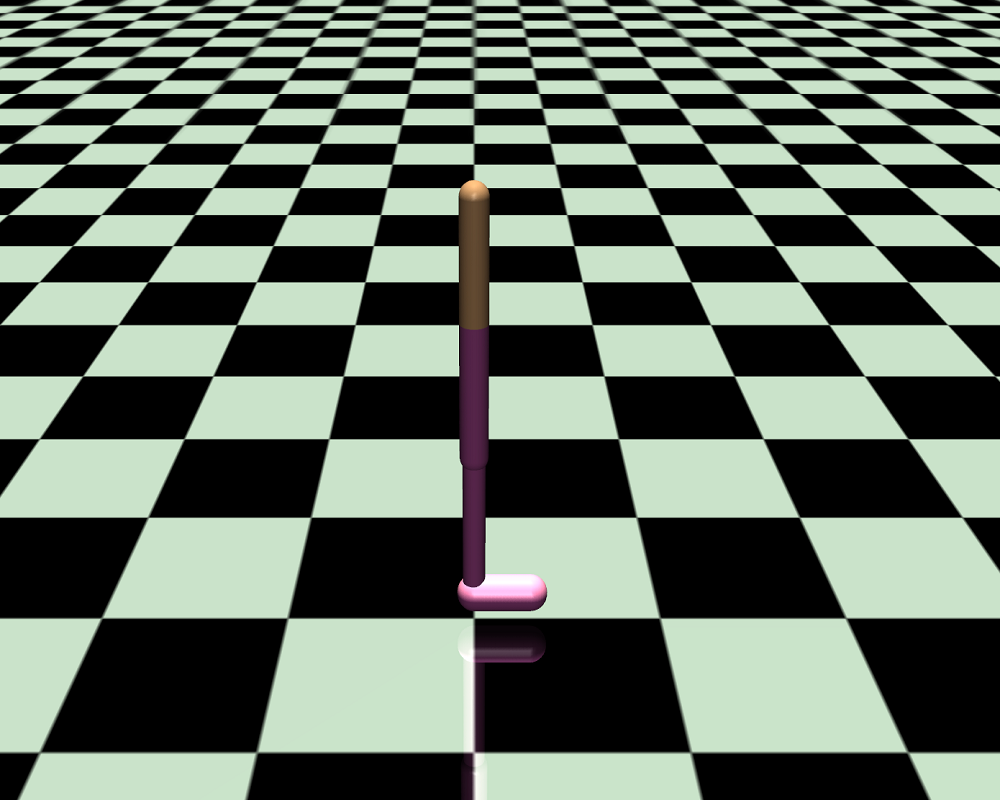}
			\end{minipage}%
		}%
		\subfigure[Hopper-v2]{
			\begin{minipage}[t]{0.5\linewidth}
				\centering
				\includegraphics[width=2.1in]{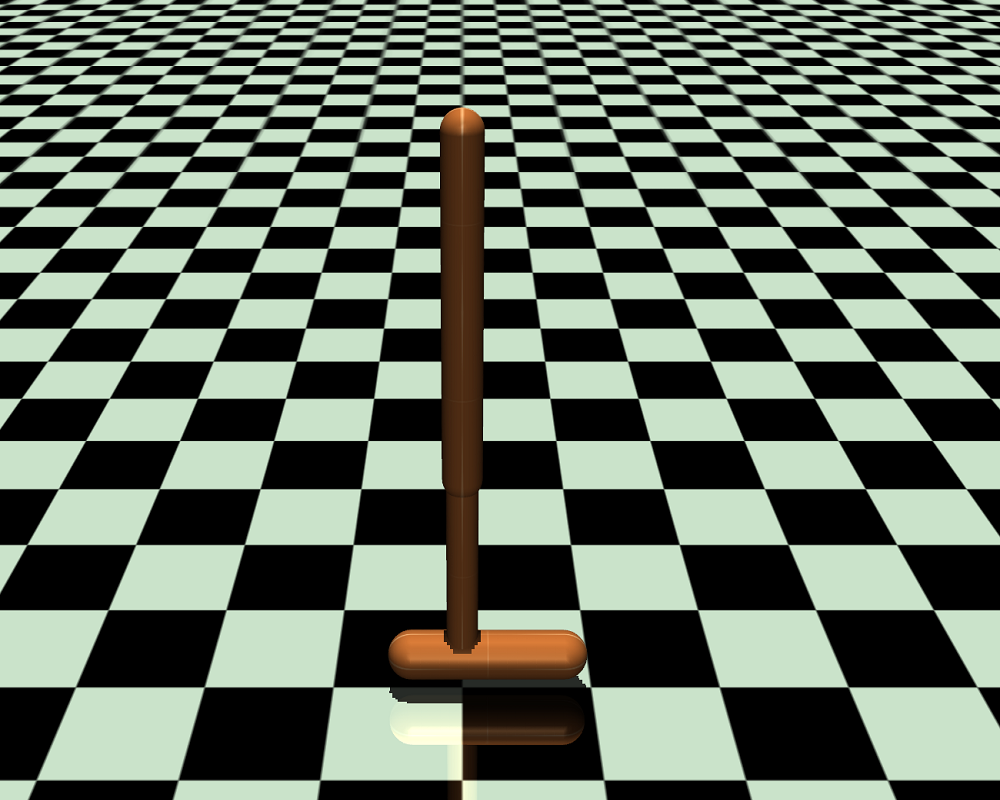}
			\end{minipage}%
		}%
	
		\subfigure[HalfCheetah-v2]{
			\begin{minipage}[t]{0.5\linewidth}
				\centering
				\includegraphics[width=2.1in]{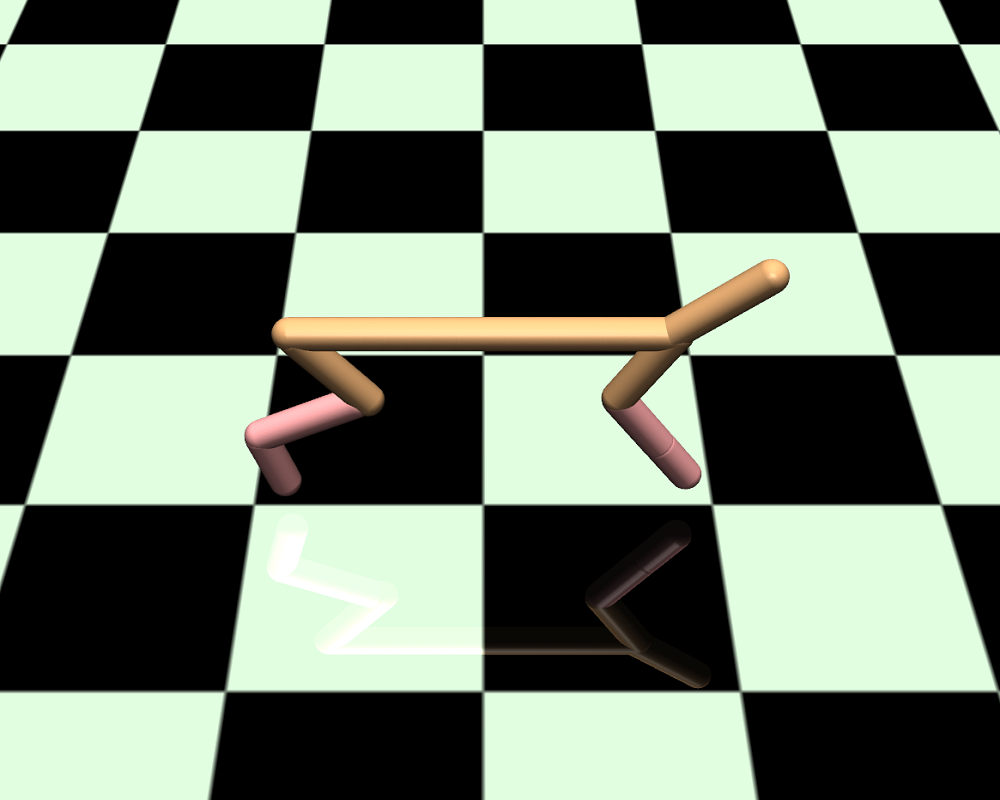}
			\end{minipage}%
		}%
		\subfigure[Swimmer-v2]{
			\begin{minipage}[t]{0.5\linewidth}
				\centering
				\includegraphics[width=2.1in]{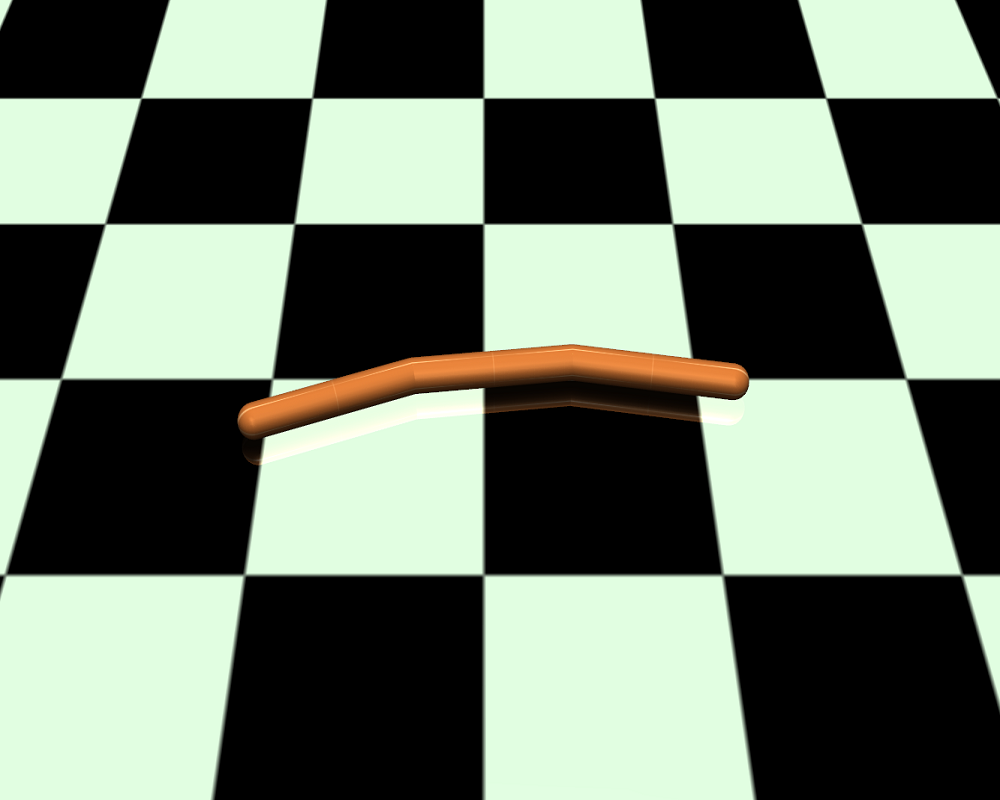}
			\end{minipage}%
		}%
		\centering
		\caption{Four continuous control tasks in Mujoco-based environments}	
		\label{fig:mujoco}
	\end{spacing}
\end{figure*}

We evaluated the performance of RIM on four continuous control tasks in Mujoco \cite{todorov2012mujoco} hosted through the OpenAI gym \cite{brockman2016openai}, which are widely used in the evaluation of reinforcement learning continuous domains \cite{duan2016benchmarking} \cite{henderson2018deep} \cite{islam2017reproducibility}. These tasks are shown in Figure \ref{fig:mujoco}. 

Baselines for comparison include Evolutionary Reinforcement Learning (ERL)  algorithm \cite{khadka2018evolution}, DDPG, and a standard evolutionary algorithm (EA). ERL is a state of the art algorithm in the Mujoco benchmarks; DDPG is considered as one of the best reinforcement learning algorithms and is widely used in a variety of continuous control tasks.

The reinforcement learning parameters of RIM are consistent with those of ERL and DDPG. The crossover and mutation probability is 0 and 0.01 respectively in population of RIM, ERL and EA. In the comparative experiments, RIM uses soft update to recruit policies from population. All actor networks and critic networks have two hidden layers, with rectified linear units (ReLU) between hidden layers. The last layer of actor networks is linked to a tanh unit. We use Adam \cite{kingma2014adam} to update all network parameters. The learning rates of the gradient policy network and the critic network are $10^{-4}$ and $10^{-3}$, respectively. The number of individuals in the population is $10$. Imitation learning is performed every 10 generations. The loss function of the imitation learning is least absolute error (L1), and the learning rate is $10^{-3}$. The sample batch size of reinforcement learning and imitation learning are 128 and 32, respectively.

The performance of the algorithms involved in the comparison is tested using different random seeds and is recorded every 10,000 frames. When testing, the average of five test results is recorded as the performance at this time. For DDPG, the actor's exploration noise was removed during the test. For EA, ERL and RIM, the performance of the best individuals in their population are recorded as the final performance. The scores of these algorithms are the rewards returned by environments, and the corresponding number of frames is the cumulative number of frames in which an algorithm interacts with the environment as a whole. These recording methods are consistent with previous comparative experiments \cite{khadka2018evolution}.

\subsection{Comparison}

\begin{figure*}[!htb]
	\begin{spacing}{0.7}	
		\centering
		\subfigure[Walker2d-v2]{
			\begin{minipage}[t]{0.5\linewidth}
				\centering
				\includegraphics[width=2.55in]{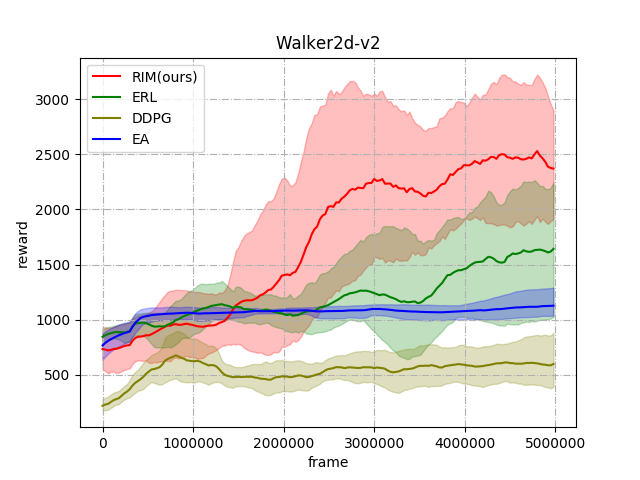}
			\end{minipage}%
		}%
		\subfigure[Hopper-v2]{
			\begin{minipage}[t]{0.5\linewidth}
				\centering
				\includegraphics[width=2.55in]{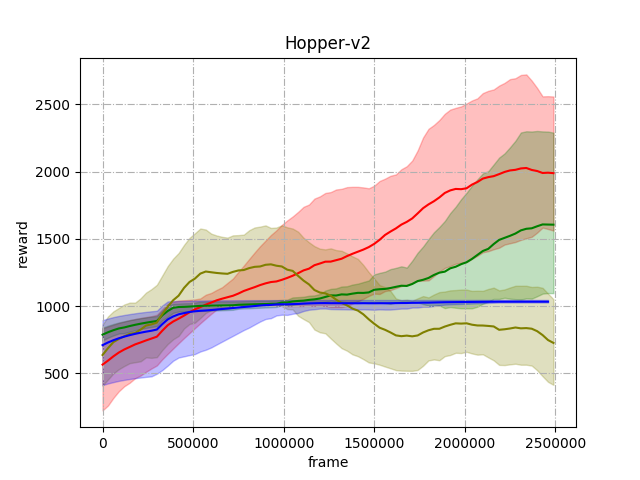}
			\end{minipage}%
		}%
		
		\subfigure[HalfCheetah-v2]{
			\begin{minipage}[t]{0.5\linewidth}
				\centering
				\includegraphics[width=2.45in]{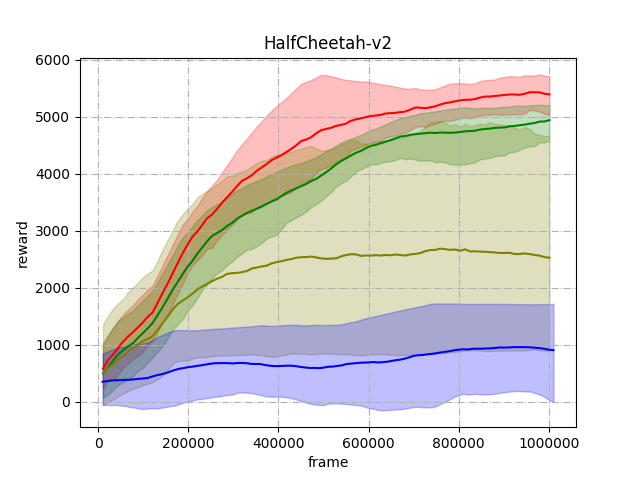}
			\end{minipage}%
		}%
		\subfigure[Swimmer-v2]{
			\begin{minipage}[t]{0.5\linewidth}
				\centering
				\includegraphics[width=2.45in]{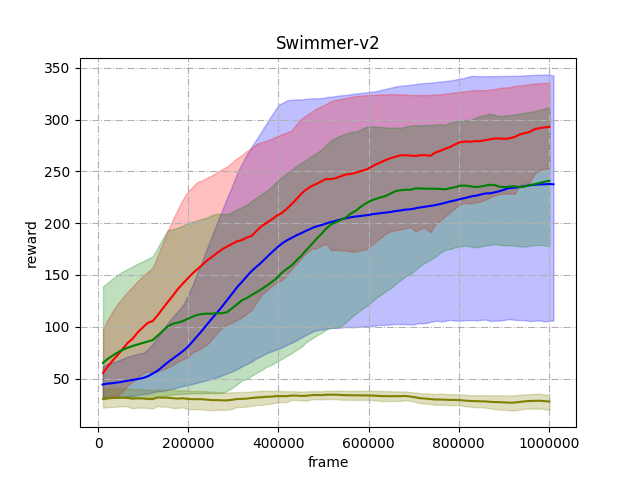}
			\end{minipage}%
		}%
		\centering
		\caption{Learning curves of EA, DDPG, ERL and RIM on 4 Mujoco-based continuous control benchmarks}	
		\label{fig:main}
	\end{spacing}
\end{figure*}

The results in Table \ref{tab:main} and Figure \ref{fig:main} show the final performance and learning curves of the four algorithms on Mujoco-based continuous control benchmarks. Each algorithm is trained for 5M frames in Walker2d-v2, 2.5M frames in Hopper-v2, and 1M frames in HalfCheetah-v2 and Swimmer-v2.  Table \ref{tab:main} presents the maximum (Max) reward obtained by each algorithm in the whole learning process, as well as the average (Mean), median (Median) and standard deviation (Std.) of 5 test results of each algorithm under different random seeds. The best statistical results have been bolded in each task. Figure \ref{fig:main} presents the average proformance and error bars, which have been smoothed using a window of size 10.

\begin{table*}  
	\begin{spacing}{1}	
		\caption{Final performance of EA, DDPG, ERL and RIM on 4 Mujoco-based continuous control benchmarks}  
		\centering
		\begin{tabular*}{12.5cm}{cccccc}
			\hline  
			\multirow{2}{*}{}&\multirow{2}{*}{}& \multirow{2}{*}{Walker2d-v2}& \multirow{2}{*}{Hopper-v2}& \multirow{2}{*}{Halfcheetah-v2}& \multirow{2}{*}{Swimmer-v2}\\\\
			\hline
			\multirow{4}{*}{EA}&{Max}&{1339.11}&{1051.19}&{1755.27}&{\textbf{356.01}}\\
			~&{Mean}&{1143.68}&{1032.92}&{920.60}&{239.04}\\
			~&{Median}&{1073.38}&{1027.82}&{864.62}&{\textbf{301.93}}\\
			~&{Std.}&{\textbf{11.37\%}}&{\textbf{0.08\%}}&{65.49\%}&{43.21\%}\\
			\hline
			\multirow{4}{*}{DDPG}&{Max}&{3184.08}&{\textbf{3575.19}}&{6012.01}&{55.32}\\
			~&{Mean}&{543.94}&{484.29}&{2601.53}&{26.78}\\
			~&{Median}&{492.85}&{590.66}&{2796.43}&{29.25}\\
			~&{Std.}&{39.77\%}&{67.12\%}&{60.66\%}&{27.86\%}\\
			\hline
			\multirow{4}{*}{ERL}&{Max}&{4472.21}&{2392.49}&{5597.36}&{332.69}\\
			~&{Mean}&{1375.24}&{1780.90}&{5098.46}&{231.38}\\
			~&{Median}&{1476.77}&{1824.88}&{5064.25}&{222.56}\\
			~&{Std.}&{23.51\%}&{30.23\%}&{\textbf{5.54\%}}&{21.37\%}\\
			\hline
			\multirow{4}{*}{RIM(ours)}&{Max}&{\textbf{5532.66}}&{3439.74}&{\textbf{6151.81}}&{343.62}\\
			~&{Mean}&{\textbf{2852.08}}&{\textbf{2385.05}}&{\textbf{5399.68}}&{\textbf{297.43}}\\
			~&{Median}&{\textbf{3242.61}}&{\textbf{2386.84}}&{\textbf{5367.22}}&{287.49}\\
			~&{Std.}&{26.66\%}&{19.66\%}&{6.94\%}&{\textbf{9.78\%}}\\
			\hline
		\end{tabular*}  	
		\label{tab:main}
	\end{spacing}
	
\end{table*}

Table \ref{tab:main} shows that the average performance of RIM exceeds the previous methods in all environments, and the maximum reward obtained in the whole learning process and median performance exceed the previous methods in most environments. Both RIM and ERL have higher variances on different random seeds because the components, such as EA and DDPG, have high variances. However, the standard deviation of RIM performance is much lower than ERL in Hopper-v2 and Swimmer-v2.

The experimental results in Figure \ref{fig:main} show that RIM can learn better than the previous algorithms in four tasks. In Walker2d-v2 and Hopper-v2 environments, the superiority of RIM is more significant, because the traditional off-policy reinforcement learning method cannot explore efficiently in such environments. EA has a stagnation period in the evolution process. Off-policy reinforcement learning individuals in ERL can help the evolutionary population to break through the stagnation period, but this often requires superior individuals in the population to generate a large amount of experiences. The recruitment mechanism of RIM enables outstanding individuals of EA to directly guide reinforcement learning individuals, instead of guiding them by generating experience. Therefore, RIM can break through the stagnation period earlier than ERL, which allows RIM to learn faster.

We reimplemented the EA, DDPG and ERL algorithms. The ERL results we presented in Walker2d and Hopper environments are consistent with Khadka et al. \cite{khadka2018evolution}. The results in the Halfcheetah environment are lower than Khadka et al., but close to those reported by Pourchot et al. \cite{pourchot2018cem}. In addition, we found that in the Swimmer environment, the performance of EA, ERL and RIM is unstable. In the best case, they can reach 330 or more, but in the worst case, they perform less than 250, even less than 150 in EA.

\subsection{Component performance}

\begin{figure*}[!htb]
	\begin{spacing}{0.7}	
		\centering
		\subfigure[Walker2d-v2]{
			\begin{minipage}[t]{0.5\linewidth}
				\centering
				\includegraphics[width=2.6in]{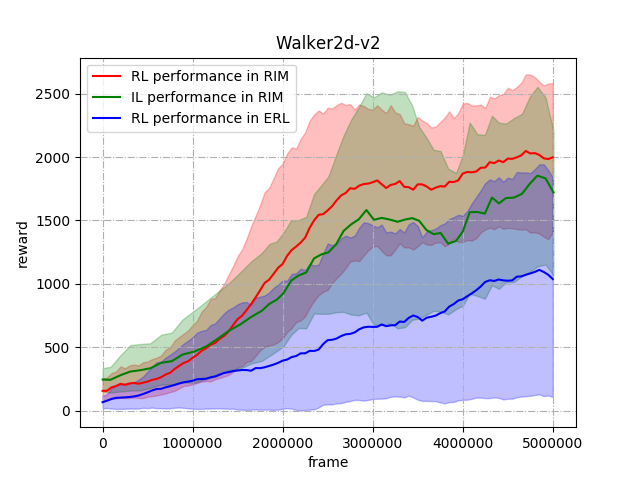}
			\end{minipage}%
		}%
		\subfigure[Hopper-v2]{
			\begin{minipage}[t]{0.5\linewidth}
				\centering
				\includegraphics[width=2.65in]{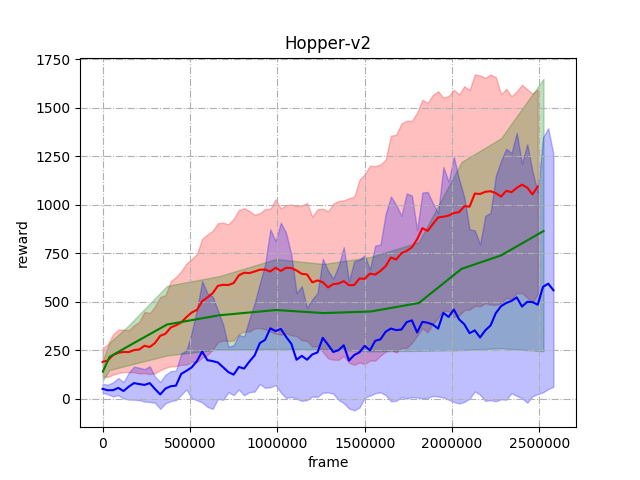}
			\end{minipage}%
		}%
		\centering
		\caption{Performance of components of different algorithms}	
		\label{fig:components}
	\end{spacing}
\end{figure*}

Figure \ref{fig:components} shows performance of the three components in Walker2d and Hopper environments: RL components of RIM, imitation learning (IL) actors in RIM, and RL agent in ERL. The performance of RL agent in RIM is better than RL agent in ERL, which means that the recruitment mechanism can accelerate the learning process of RL actors. The performance of individuals trained by imitation learning in RIM improves as the performance of the RL components improves, which illustrates the effectiveness of off-policy imitation learning. 

The externally injected individuals accepted by RIM's population are imitation learning individuals, while the externally injected individuals accepted by the ERL's population are RL actors. In Figure \ref{fig:components}, the performance of imitation learner in RIM outperforms RL agent in ERL, suggesting that externally injected individuals of the population of RIM are better. This can directly explain why the learning speed of RIM is better than that of ERL.

Table \ref{tab:selection-rate} presents the selection rate of actors trained through imitation learning during evolution.\begin{table}[!h]
	\caption{Selection rate for $\pi_{wt}$ after training}  
	\centering
	\begin{tabular*}{6cm}{ccc}
		\hline  
		{}& {Selected}& {Discarded}\\
		\hline
		{Walker2d-v2}&  {59.87$\%$}& {40.13$\%$}\\
		{Hopper-v2}&  {30.92$\%$}& {69.08$\%$}\\
		{HalfCheetah-v2}&  {70.37$\%$}& {29.63$\%$}\\
		{Swimmer-v2}& {24.44$\%$}& {75.56$\%$}\\
		\hline
	\end{tabular*}  	
	\label{tab:selection-rate}
\end{table} When the population selects excellent individuals, $\pi_{wt}$ with higher performance has higher probability to be selected. On the contrary, $\pi_{wt}$ with lower performance will have a higher probability of being discarded. The RL components performs better in Walker2d and Halfcheetah, so the selected probability of well-learned $\pi_{wt}$ is higher. In Hopper environment, the RL components performs poorly before breaking through the long stagnation period. In the Swimmer environment, the RL components cannot play a critical role in the whole learning process, which can be seen from curve of DDPG in Figure \ref{fig:main}. These reasons cause $\pi_{wt}$ cannot imitate a good policy, so the probability that $\pi_{wt}$ is selected is lower. On the whole, individuals performing imitation learning can be selected in each environment, indicating that the policy learned by imitation learning can accelerate the evolution of the population.

\subsection{Soft update in recruitment}  

\begin{figure*}[t]
	\begin{spacing}{0.8}	
		\centering
		\subfigure[HalfCheetah-v2]{
			\begin{minipage}[t]{0.5\linewidth}
				\centering
				\includegraphics[width=2.5in]{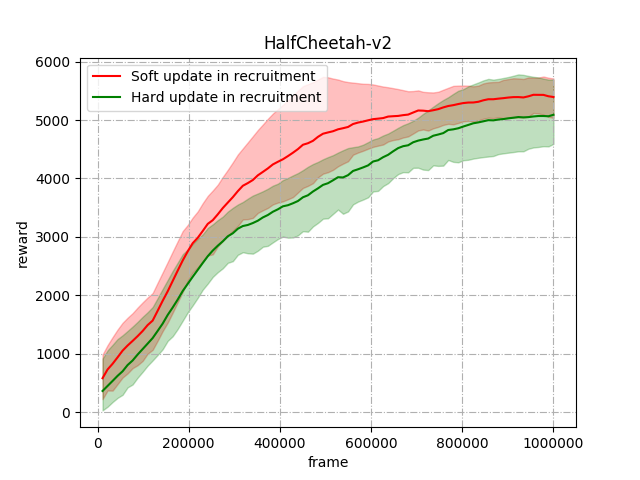}
			\end{minipage}%
		}%
		\subfigure[Hopper-v2]{
			\begin{minipage}[t]{0.5\linewidth}
				\centering
				\includegraphics[width=2.5in]{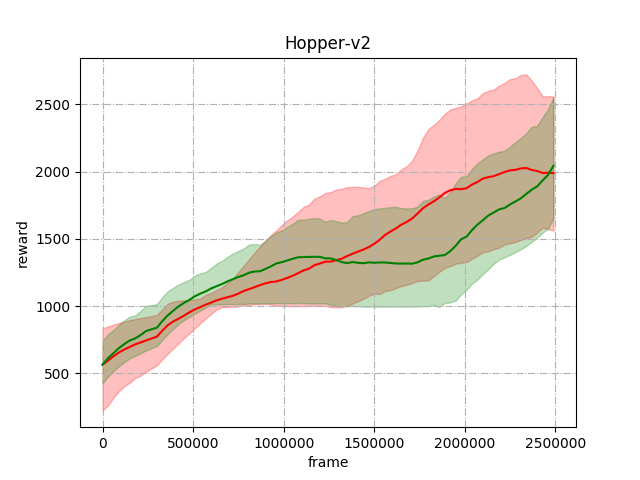}
			\end{minipage}%
		}%
		\centering
		\caption{Comparison curves between soft update recruitment and hard update recruitment}	
		\label{fig:soft-hard}
	\end{spacing}
\end{figure*}

We tested the performance of RIM using hard update, and compared it with the original RIM. As shown in Figure \ref{fig:soft-hard}, RIM using soft update recruitment performed slightly better than RIM using hard update recruitment. Because in the process of guiding the gradient policy network learning, the soft update recruitment policy network can provide more stable $y^{(i)}$ when calculating equation (4), which ensures better consistency for learning process. However, recruitment using hard update does not make the policy gradient network unable to learn. In fact, as the population iterates, the changes of parameters between generations are usually small. Therefore, parameter changes of the recruitment policy network are small, which can also bring a certain degree of consistency to the learning process.

\subsection{Ablation experiments} 
\begin{figure*}[]
	\begin{spacing}{0.7}	
		\centering
		\subfigure[Walker2d-v2]{
			\begin{minipage}[t]{0.5\linewidth}
				\centering
				\includegraphics[width=2.55in]{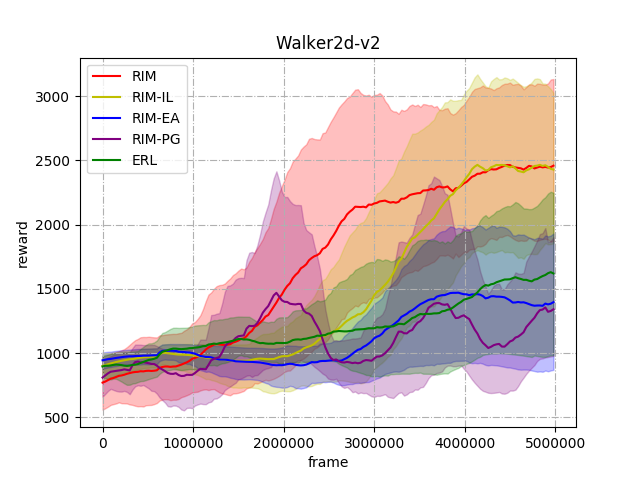}
			\end{minipage}%
		}%
		\subfigure[Hopper-v2]{
			\begin{minipage}[t]{0.5\linewidth}
				\centering
				\includegraphics[width=2.55in]{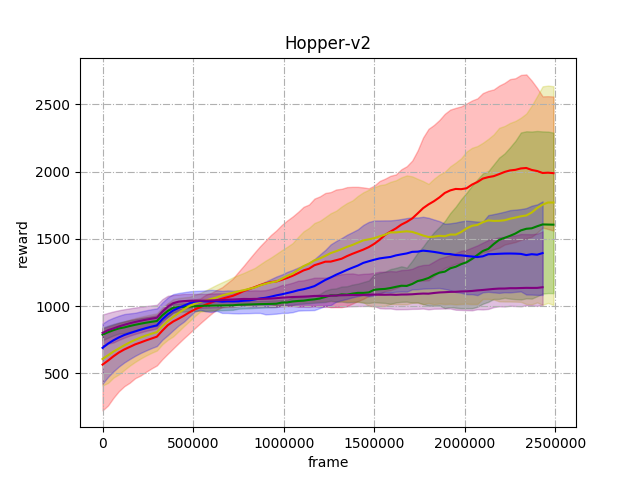}
			\end{minipage}%
		}%
		
		\subfigure[HalfCheetah-v2]{
			\begin{minipage}[t]{0.5\linewidth}
				\centering
				\includegraphics[width=2.45in]{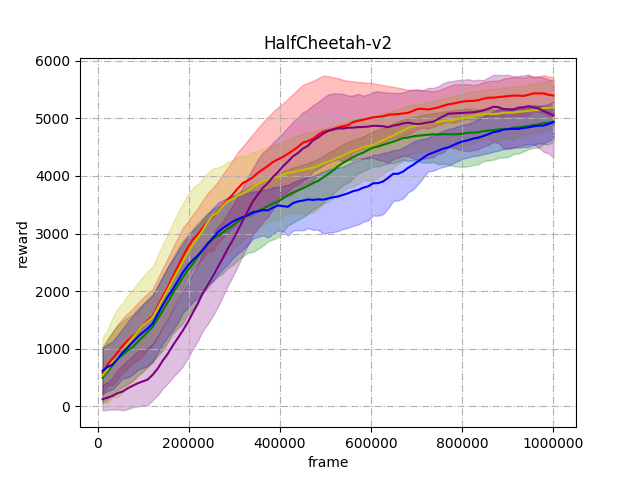}
			\end{minipage}%
		}%
		\subfigure[Swimmer-v2]{
			\begin{minipage}[t]{0.5\linewidth}
				\centering
				\includegraphics[width=2.45in]{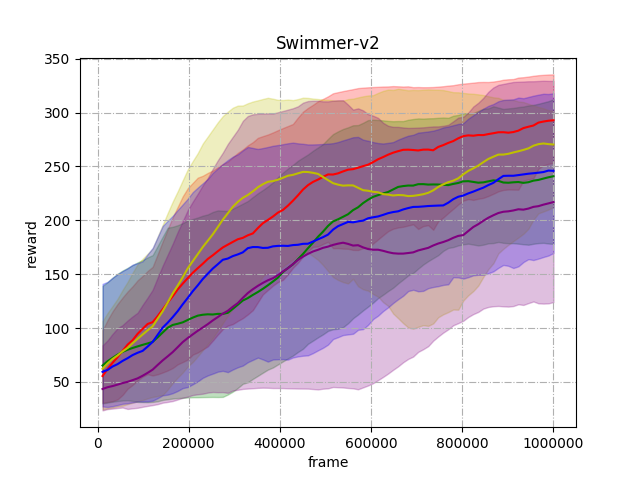}
			\end{minipage}%
		}%
		\centering
		\caption{Learning curves of ERL, RIM and variants of RIM on 4 Mujoco-based continuous control benchmarks}	
		\label{fig:rim-}
	\end{spacing}
\end{figure*}

In order to further prove the effectiveness of the RIM components, we performed ablation experiments on RIM. We tested the performance of RIM without offline imitation learning (RIM-IL), RIM without using the recruitment network to estimate $Q'$ (RIM-EA), and RIM without using the gradient policy network to estimate $Q'$ (RIM-PG). We compare the test performance of these three RIM variants with RIM as well as ERL. The comparison results are presented in Table \ref{tab:rim-} and Figure \ref{fig:rim-}.

\begin{table*}  
	\begin{spacing}{1}	
		\caption{Final performance of ERL, RIM and variants of RIM on 4 Mujoco-based continuous control benchmarks}  
		\centering
		\begin{tabular*}{12.5cm}{cccccc}
			\hline  
			\multirow{2}{*}{}&\multirow{2}{*}{}& \multirow{2}{*}{Walker2d-v2}& \multirow{2}{*}{Hopper-v2}& \multirow{2}{*}{Halfcheetah-v2}& \multirow{2}{*}{Swimmer-v2}\\\\
			\hline
			\multirow{3}{*}{RIM}&{Mean}&{\textbf{2852.08}}&{\textbf{2385.05}}&{\textbf{5399.68}}&{\textbf{297.43}}\\
			~&{Median}&{\textbf{3242.61}}&{\textbf{2386.84}}&{\textbf{5367.22}}&{\textbf{287.49}}\\
			~&{Std.}&{26.66\%}&{19.66\%}&{6.94\%}&{9.78\%}\\
			\hline
			\multirow{3}{*}{RIM-IL}&{Mean}&{2075.33}&{1574.41}&{5229.43}&{269.68}\\
			~&{Median}&{1937.64}&{1095.01}&{5253.77}&{267.89}\\
			~&{Std.}&{26.43\%}&{56.14\%}&{7.5\%}&{\textbf{9.45\%}}\\
			\hline
			\multirow{3}{*}{RIM-EA}&{Mean}&{1948.16}&{1753.66}&{5001.21}&{258.96}\\
			~&{Median}&{2018.73}&{1749.26}&{4923.92}&{261.63}\\
			~&{Std.}&{49.10\%}&{29.14\%}&{7.67\%}&{20.63\%}\\
			\hline
			\multirow{3}{*}{RIM-PG}&{Mean}&{1777.30}&{1141.54}&{5142.67}&{222.60}\\
			~&{Median}&{1208.66}&{1078.56}&{5197.68}&{209.84}\\
			~&{Std.}&{54.97\%}&{\textbf{15.23\%}}&{13.96\%}&{32.61\%}\\
			\hline
			\multirow{3}{*}{ERL}&{Mean}&{1375.24}&{1780.90}&{5098.46}&{231.38}\\
			~&{Median}&{1476.77}&{1824.88}&{5064.25}&{222.56}\\
			~&{Std.}&{\textbf{23.51\%}}&{30.23\%}&{\textbf{5.54\%}}&{21.37\%}\\
			\hline
			
		\end{tabular*}  	
		\label{tab:rim-}
	\end{spacing}
	
\end{table*}

According to the results in Table \ref{tab:rim-} and Figure \ref{fig:rim-}, RIM outperforms RIM-IL, which shows that imitation learning works. We believe that imitation learning can inject individuals with dual-policy RL agent's behavior patterns into the population, thereby accelerating the evolution of the population. Performance of RIM is better than that of RIM-EA and RIM-PG, which shows that the dual-policy learning mode can learn better policy. This stems from the fact that the dual-policy can better estimate the $Q'$ value, as revealed in Theorem 1. 

\begin{figure*}[t]
	\begin{spacing}{0.8}	
		\centering
		\subfigure[Wakler2d-v2]{
			\begin{minipage}[t]{0.5\linewidth}
				\centering
				\includegraphics[width=2.5in]{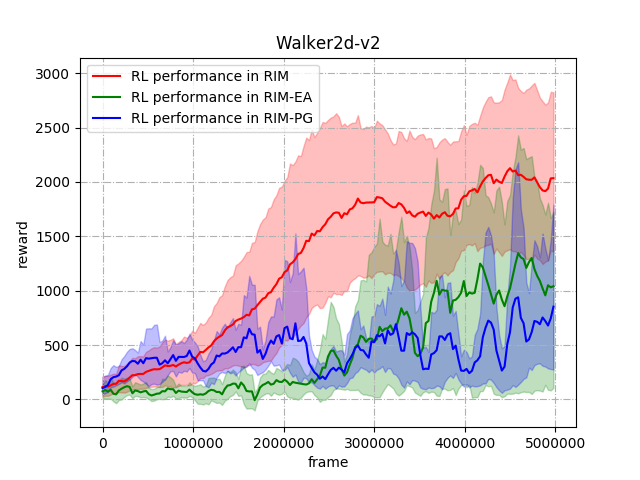}
			\end{minipage}%
		}%
		\subfigure[Hopper-v2]{
			\begin{minipage}[t]{0.5\linewidth}
				\centering
				\includegraphics[width=2.5in]{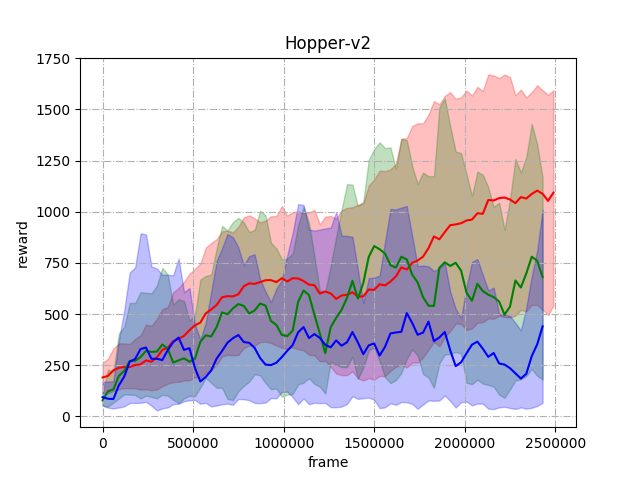}
			\end{minipage}%
		}%
		\centering
		\caption{Learning curves of RL components in RIM, RIM-PG, and RIM-EA}	
		\label{fig:dual-policy}
	\end{spacing}
\end{figure*}

To further illustrate the effectiveness of the dual-policy learning model in evolutionary reinforcement learning algorithms, we compared learning curves of RL components in RIM, RIM-EA, and RIM-PG algorithms. These curves are shown in Figure \ref{fig:dual-policy}. RIM's RL policy is better than RIM-EA, which indicates that RIM estimates the $Q'$ value better than DDPG. RIM-PG's RL components performs poorly. This results from the fact that the $Q'$ estimated by the recruitment network is not accurate in Walker2d-v2 and Hopper-v2.

\section{Conclusion and Future Work}

In this paper, we develop a dual-actors and single critic RL agent, which can be well combined with evolutionary algorithms. Based on this, we propose RIM for evolutionary reinforcement learning. This method not only outperforms the previous evolutionary and off-policy reinforcement learning algorithms, but also exceeds ERL both in overall performance and component performance. Finally, we experimentally demonstrated that the recruitment using soft update enables RL agent to learn faster than that using hard update.

Future work can be explored from the following three aspects:
\begin{itemize}
	\item RIM uses a standard evolutionary algorithm to iterate populations. Other more efficient evolutionary algorithms may provide immediate improvements to RIM's performance, such as Natural Evolution Strategy (NES) \cite{wierstra2008natural}, NeuroEvolution of Augmenting Topologies (NEAT) \cite{stanley2002evolving}.
	\item The dual-actors and single critic RL agent in RIM is extended based on DDPG, but in fact, any off-policy actor-critic deep reinforcement learning method can apply this structure, such as Soft Actor-Critic (SAC) \cite{haarnoja2018soft}, Twin Delayed Deep Deterministic policy gradient algorithm (TD3) \cite{fujimoto2018addressing}. Since these methods are superior to DDPG in most cases, applying these methods to the dual-actors and single critic RL agent may improve RIM's performance.
	\item In RIM's experimental settings, at each time 100,000 reinforcement learning gradient updates are performed, then 30,000 imitation learning gradient updates are required. Parallel methods can be employed to reduce time for training models.
\end{itemize}

\section*{Acknowledgement}

This work was supported by the National Key R\&D Program of China under Grant No. 2017YFB1003103; the National Natural Science Foundation of China under Grant Nos. 61300049, 61763003; and the Natural Science Research Foundation of Jilin Province of China under Grant Nos. 20180101053JC, 20190201193JC.

\section*{References}

\bibliography{mybibfile}

\end{document}